%% file: main.tex
\documentclass[letterpaper, 10 pt, conference]{ieeeconf}

\IEEEoverridecommandlockouts                              %

\usepackage{xspace}
\usepackage{xcolor}

\newcommand{\transpose}{\mbox{${}^{\text{T}}$}}

\newcommand{\acronym}{CREST\xspace}
\usepackage{mwe} %
\usepackage{bm}
\usepackage{amsfonts} %
\usepackage{subcaption}
\usepackage{graphicx}       %
\graphicspath{{figures/}}   %
\usepackage{amsmath}
\usepackage[ruled,vlined]{algorithm2e}
\usepackage{balance}
\usepackage{makecell}
\usepackage{hyperref}
\usepackage[noadjust]{cite}

\usepackage[font=small,labelfont=bf]{caption}

\title{Causal Reasoning in Simulation for Structure and Transfer Learning \\ of Robot Manipulation Policies}

\author{
  Tabitha Edith Lee, Jialiang~(Alan)~Zhao, Amrita S.~Sawhney, Siddharth~Girdhar, and Oliver Kroemer
  \thanks{All authors are affiliated with the Robotics Institute, Carnegie Mellon University, Pittsburgh PA 15123, USA. {\tt\small tabithalee@cmu.edu} 
  Supplementary materials: \url{https://sites.google.com/view/crest-causal-struct-xfer-manip}
  }
}

\begin{document}
\maketitle

\input{content/0_abstract}

\input{content/1_introduction}

\input{content/2_related_works}

\input{content/3_formulation}

\input{content/4_causal_relevance}

\input{content/5_policy_learning}

\input{content/6_experiments}

\input{content/7_conclusion}

\input{content/8_acknowledgments}

\clearpage
\balance
\bibliographystyle{IEEEtran.bst}
\bibliography{IEEEabrv.bib, references.bib}

\clearpage
\input{content/9_appendix}

\end{document}

%% file: content/0_abstract.tex
\begin{abstract}
We present \acronym, an approach for causal reasoning in simulation to learn the relevant state space for a robot manipulation policy.
Our approach conducts interventions using internal models, which are simulations with approximate dynamics and simplified assumptions.
These interventions elicit the structure between the state and action spaces, enabling construction of neural network policies with only relevant states as input.
These policies are pretrained using the internal model with domain randomization over the relevant states.
The policy network weights are then transferred to the target domain (e.g., the real world) for fine tuning.
We perform extensive policy transfer experiments in simulation for two representative manipulation tasks: block stacking and crate opening.
Our policies are shown to be more robust to domain shifts, more sample efficient to learn, and scale to more complex settings with larger state spaces.
We also show improved zero-shot sim-to-real transfer of our policies for the block stacking task. %
\end{abstract}

%% file: content/1_introduction.tex
\section{INTRODUCTION}
\label{sec:intro}
Real-world environments, such as homes and restaurants, often contain a large number of objects that a robot can manipulate.
However, usually only a small set of objects and state variables are actually relevant for performing a given manipulation task.
The capability of reasoning about what aspects of the state space are relevant for the task would lead to more efficient learning and greater skill versatility.

Current approaches to learning versatile manipulation skills often utilize simulation-to-real (sim-to-real) transfer learning~\cite{peng2018sim, bousmalis2018using, kroemer2019review, zhao2020sim}, wherein the skill is learned in a simulation and then deployed and fine-tuned (if feasible) on the real robot.  
Sim-to-real learning is often combined with domain randomization (DR)~\cite{tobin2017domain}, which involves training the skill on a wide range of task instances in simulation such that the resulting skill is more robust and generalizes across task variations.
However, for scenes with distractor objects, the policy still takes the irrelevant state features as inputs.
Larger domain shifts in the irrelevant features can therefore still be detrimental to the performance of the skill policy.
Rather than relying only on DR, the robot can use model-based reasoning to identify the ``structure'' of a policy --- the interplay between relevant state inputs and control outputs.

In this paper, we propose using causal reasoning to improve sim-to-real transfer through conducting interventions in simulation to determine which state variables are relevant for the successful execution of a task using a given controller.
We refer to the resulting algorithm as \textbf{CREST}: \textbf{C}ausal \textbf{R}easoning for \textbf{E}fficient \textbf{S}tructure \textbf{T}ransfer. %
The relevant state variables from \acronym are used to construct policies that encode the causal structure of the manipulation task.
These policies are initially trained in simulation using domain randomization over only the states that have explicitly been determined as relevant.
Moreover, policies that only use relevant state variables require significantly fewer parameters and, by construction, are robust to distribution shifts in irrelevant state spaces.
In this manner, our approach produces lightweight policies that are designed for efficient online adaptation to unforeseen distribution shifts that may occur when bridging the sim-to-real gap.
This contrasts with existing sim-to-real approaches that train large policies over enormous state spaces, where the costs for achieving robust zero-shot transfer may be intractable. %

Our proposed approach was successfully evaluated on block stacking and crate opening tasks.
Although our method is intended for sim-to-real transfer, we primarily conduct our experiments by transferring to task simulations in NVIDIA Isaac Gym~\cite{liang2018gpu}, a high-fidelity physics simulator, as a proxy for real systems.
This is necessary to experimentally evaluate distributions shifts that would be intractable to evaluate (but nonetheless feasible) in practical manipulation scenarios.
Additionally, we validate our approach for zero-shot, sim-to-real performance for the block stacking task.

The contributions of our work are as follows.
\begin{itemize}
    \item We propose \acronym, an algorithm that uses causal reasoning in simulation to identify the relevant input state variables for generalizing manipulation policies.
    \item We propose two neural network architectures that are constructed using the causal information from \acronym. 
    \item We conduct rigorous transfer learning experiments to demonstrate these policies generalize across task scenarios, scale in relevant state complexity, and are robust to distribution shifts.
    \item We propose \acronym as one approach to a broader methodology of structure-based transfer learning from simulation as a new paradigm for sim-to-real robot learning, i.e., \textit{structural} sim-to-real. %
\end{itemize}

%% file: content/2_related_works.tex
\section{RELATED WORKS}
\label{sec:litreview}

Our work relates to research areas in robotics and machine learning for structure-based learning, causality, attention, and sim-to-real transfer of higher-level policies.

\textbf{Structure-based learning, causality, and attention.}
Causal reasoning for structure, transfer, and reinforcement learning is an emerging area of research~\cite{zhang2018learning, scholkopf2019causality}, having been demonstrated for transferring multi-armed bandits policies~\cite{zhang2017transfer}, examining distribution shifts via imitation learning~\cite{de2019causal}, modeling physical interactions from videos~\cite{li2020causal}, and clustering causal factors~\cite{sontakke2020causal}.
Causality reasons about the data generation process with respect to an underlying model~\cite{pearl2009causality}, which our CREST policies encode.
The motivation of our approach to transfer learned causal structure is similar to that described by~\cite{ahmed2020causalworld}. In our work, we represent this structure through the policy inputs, and we demonstrate the approach achieves sim-to-real transfer.

Our view of policy structure can be seen as an explicitly encoded form of state space attention, achieved via construction of policies using only relevant inputs.
We are primarily concerned with object feature states, which enable significantly smaller neural networks to be constructed under our assumptions.
As a comparison, \cite{devin2018deep} learns implicit attention to task-relevant objects to generalize manipulation skills.
This approach requires a few example trajectories to be provided and uses a vision-based state representation, making explicit reasoning about states more challenging.

Similar to~\cite{yu2017preparing}, our approach can also generalize policies to unforeseen dynamic distribution shifts; ours does so primarily through more efficient fine-tuning.

Our work is also similar in spirit to the work of Nouri and Littman~\cite{nouri2010dimension} in terms of achieving dimensionality reduction for reinforcement learning.
Whereas our work seeks to reduce the dimensionality for the state's possible influence on the task policy under different contexts, they instead demonstrate dimensionality reduction in the action space.
Kolter and Ng~\cite{kolter2009regularization} as well as Parr et al.~\cite{parr2008analysis} approximate the value function by learning the relevant basis functions.

\textbf{Sim-to-real and higher-level policies.} Our work is a form of simulation-to-reality transfer of controllers~\cite{tobin2017domain, peng2018sim, tan2018sim, molchanov2019sim}.
Unlike in the typical sim-to-real paradigm, our policies are not required to transfer zero-shot.
Rather, the dimensionality reduction afforded by transferring the relevant state space with \acronym enables more efficient online adaptation.
Unlike~\cite{chebotar2019closing}, we do not transfer any target samples back to the internal model. %
Similar to the problem settings of \cite{masson2016reinforcement, hausknecht2015deep, fan2019hybrid}, our agent predicts the best parameters for a controller. %

%% file: content/3_formulation.tex
\section{Problem Formulation}
\label{sec:preliminaries}

We formulate our problem as a multi-task reinforcement learning problem, wherein a policy $\pi$ is learned to complete a series of tasks $\mathcal{T}_i$.
Each task is modeled as a Markov decision process (MDP).
The state space $\mathcal{S}$ and action space $\mathcal{A}$ are the same across tasks.
However, each task $\mathcal{T}_i$ defines a separate initial state $s_0(c_i)$, transition function $p(s'|s,a,c_i)$, and reward function $r(a,s,c_i)$ that are parameterized by the task's context variables $c_i\in \mathcal{C}$.
We assume that the robot has an internal model $p_\text{int}(s'|s,a,c_i)$ that approximates the transition function of the target task domain $p(s'|s,a,c_i)$.%
The context variables capture object parameter variations, such as shapes, appearances, and initial states.
We assume that the context variables are always set such that the task is feasible.

To solve the tasks, the robot learns a policy $\pi(a|s,c)$ that is decomposed into two parts~\cite{deisenroth2013survey}: an upper policy $\pi(\theta|c)$ and a lower control policy $\pi_\theta(a|s)$, where $\theta\in\mathbb{R}^d$ are the controller parameters.
Transferring the parameters from the internal model to the target domain may require fine-tuning.
At the start of each task, the upper policy, which is responsible for generalizing between different task contexts, selects a set of parameters for the control policy to use throughout the task execution.
We use multilayer perceptron (MLP) neural networks for the upper policy.
In principle, the control policies can take on a variety of parameterized forms, such as motor primitives, planners, waypoint trajectories, or linear feedback controllers, depending on the task.
We assume a control policy with known preconditions is available.

Of the context variables $c$ that describe the object variations in the scene, only a subset may be relevant for the policy.
We refer to this subset of relevant variables as $\tau \subseteq c$, such that the robot can learn a policy $\pi(\theta|\tau)$ to complete the tasks.
Our goal is to determine $\tau$ through causal reasoning with the internal model, yielding policies with fewer input variables as compared to naively using all of $c$.

The set of controller parameters can be divided into individual parameters $[\theta]_j\forall j\in 1,...,d$, where $[\theta]_j$ indicates the $j$th element of vector $\theta$.
Each of these parameters may rely on a different set of relevant variables.
We can thus further divide the problem into determining a set of context variables $\tau_j$ for each policy parameter $[\theta]_j$, such that the robot can learn a partitioned task policy $\pi([\theta]_j|\tau_{j})\forall j\in 1,...,d$.

%% file: content/4_causal_relevance.tex
\section{CAUSAL STRUCTURE LEARNING}
\label{sec:causal_relevance}
    
\acronym uses an internal model of the task to learn the relevant context $\tau$ and parameter-specific mappings $\tau_j$ that define the structure of the upper policy (c.f, Sec.~\ref{sec:networks}).

\subsection{Internal Model for Causal Reasoning}
Our approach assumes an \textit{internal model}, an approximate simulation of the task, is available.
Analogous to mental models and approximate physics models~\cite{battaglia2013simulation, ha2018recurrent}, the internal model facilitates reasoning about the effects of different context variations $\Delta c$ on completing the task with policy parameters $\theta$.
Varying the relevant context parameters $\tau$ will affect the execution and outcome of the task in the internal model while irrelevant ones will not.

Given its approximate nature, the solution obtained in the internal model is not necessarily expected to transfer zero-shot to reality.
Instead, the internal model provides both an estimate of the policy structure and a task-specific initialization via network pretraining. %
Intuitively, it is easier to reason about \textit{which} variables are important for a model rather than exactly characterizing the exact model itself. %
For example, the internal model can capture that the weight of an object affects the required pushing force, but the exact details of frictional interaction may be approximated for the purposes of pretraining the policy. %
We assume that the internal model approximates the task sufficiently well and includes all context variables $c$ that vary in the target domain.
Given the existing challenges in causal representation learning~\cite{scholkopf2021toward}, we additionally assume the representation of $c$ is amenable for determining the underlying model for $\theta$ via causal interventions~\cite{pearl2009causality}.

\subsection{Causal Reasoning to Determine Relevant Contexts}
\label{sec:ContextRelevance}
At its core, \acronym uses simulation-based causal reasoning to determine the relevant context variables for the policy input. This process is divided into two phases: 1) determining the overall set of relevant variables $\tau$, and 2) determining the relevant variables $\tau_j$ for specific policy parameters $[\theta]_j$.

\textbf{Causal interventions to determine relevant variable set.}

In this phase, the relevance of a state variable is determined by posing the following question: ``If a context variable were different, would the same policy execution still complete the task successfully?''
To answer this, we first uniformly sample a context $c_i\sim p(c)$ and solve the resulting task in simulation to acquire the corresponding policy parameters $\theta_i$. In practice, the policy parameters are optimized using Relative Entropy Policy Search~\cite{peters2010relative}.
Importantly, the policy is solved for only this specific context $c_i$ (not the general policy). 

Given the solved task, we conduct interventions $\Delta c$ to determine if the policy parameters $\theta_i$ remain valid for the new context $c_i+\Delta c$.
Interventions $\Delta c$ are conducted to only alter one context variable at a time, i.e. $||\Delta c||_0=1$. If the policy subsequently failed, the intervened variable is considered relevant and thus appended to $\tau$.
If the policy succeeded, despite the intervention, then the variable is considered irrelevant and is not required for the general policy.
The resulting relevant variable set $\tau$ is sufficient for constructing Reduced MLP policies (c.f., Sec.~\ref{sec:networks}).

\textbf{Causal interventions to determine individual policy mappings.}
Reducing the set of state variables from the full set $c$ to the relevant set $\tau$ may greatly reduce the size of the policy input.
However, for some problems, each of the policy parameters may only depend on a subset of $\tau$.
Therefore, in the next phase, the individual mappings from the relevant state variables to the controller parameters are determined by posing the question: ``Does altering this relevant context variable require this policy parameter to be changed?''

We begin from the previous phase, where $c_i$ with solution $\theta_i$ is available with interventions $\Delta c$. In this phase, interventions are only applied to relevant context variables $\tau$.

For each new context $c_i+\Delta c$, the task is solved to obtain the resulting policy parameters $\theta_i+\Delta \theta$, starting the optimization from the original parameters $\theta_i$.
The optimization will often alter all of the policy parameters, i.e., all elements of $\Delta \theta$ are non-zero, and the magnitude of their changes are not reliable estimators of their importance.
Instead, a solution that minimizes the number of non-zero changes, i.e., $\min|| \Delta \theta||_0$, is obtained using search. 
This search involves setting subsets of elements in $\Delta \theta$ to zero and evaluating if the resulting policy still solves the task with context $c_i+\Delta c$.
In our experiments, we used a breadth-first search to find a solution with a minimal set of parameter changes. %
Once a subset of parameter changes has been found, the context variable intervened on in $\Delta c$ is added to the sets $\tau_j$ of relevant inputs for the policy parameters with non-zero elements in the final $\Delta \theta$. 
The output of this phase (the parameter-specific variables $\tau_j$) can then be used to learn Partitioned MLP policies (c.f., Sec.~\ref{sec:networks}). %

\subsection{CREST Evaluation}\label{sec:crest_eval}
Although we primarily use \acronym for manipulation policies, we quantify the accuracy of \acronym in a limited, application-agnostic manner.
Table~\ref{tab:causal_analysis} shows the results of \acronym on an environment that replicates the causal structure of hierarchical manipulation policies. This environment is designed so that, given a set of ground truth mappings between context variables and policy parameters, a controller with randomized structure is generated for the agent to ``manipulate'' the environment to a goal location determined by the causal structure of the problem.
These mappings were either linear or (weakly) nonlinear.

Under the assumptions of our controller and the context $c$, \acronym excels at determining whether a variable is relevant.
This is expected by construction of the underlying mathematics of this environment and represents an expected upper bound.
We choose action space dimension of 8 and 20 (larger than our transfer learning experiments, c.f., Sec.~\ref{sec:experiments}) and select the state space accordingly to permit calculation of ground truth for testing.
We also introduce action noise to test the robustness to uncertainty from interventions, leading to more variables detected.
Relatively higher false positive rates arise in higher dimensions, but the overall reduction to the relevant set of variables is nonetheless significant.

\begin{table}
    \centering
    \captionsetup{font=small}
    \caption{\acronym evaluation for a toy environment on an aggregate (``Agg.'') and mapping-specific (``Map.'') basis. Accuracy (``Acc.'') is whether all relevant states were detected. False positives (``F.P.'') are states that were incorrect detected as relevant. 100 trials are used.}
    \label{tab:causal_analysis}
    \begin{scriptsize}
    \begin{tabular}{c | c | c | c | c | c | c}
        Class &
            Dim. &
            Noise &
            \makecell{Agg. \\ Acc.} &
            \makecell{Agg. \\ F.P.} & 
            \makecell{Map. \\ Acc.} &
            \makecell{Map. \\ F.P.} \\
        \hline
            Linear & 8 & None & 
            1.00 & 0.00 & 0.98 & 0.19 \\
        \hline
            Nonlinear & 8 & None & 
            1.00 & 0.00 & 0.97 & 0.23 \\
        \hline
            Linear & 20 & None & 
            1.00 & 0.00 & 0.99 & 0.53 \\
        \hline
            Nonlinear & 20 & None & 
            1.00 & 0.00 & 0.97 & 0.24 \\
        \hline
            Linear & 8 & Limited & 
            1.00 & 0.23 & 0.99 & 0.32 \\
        \hline
            Nonlinear & 8 & Limited & 
            1.00 & 0.13 & 0.95 & 0.22 \\
        \hline
            Linear & 20 & Limited & 
            1.00 & 0.22 & 0.98 & 0.50 \\
        \hline
            Nonlinear & 20 & Limited & 
            1.00 & 0.12 & 0.96 & 0.23 \\
    \end{tabular}
    \end{scriptsize}
    \vspace{-5mm}
\end{table}

%% file: content/5_policy_learning.tex
\section{POLICY LEARNING AND TRANSFER}
\label{sec:policy_learning}

Given the relevant context from \acronym, the structure and transfer learning pipeline of our work begins through 1) constructing the appropriate policy; 2) pretraining using the internal model; and 3) fine-tuning in the target setting.

\subsection{Policy Architectures}\label{sec:networks}

\begin{figure}
    \centering
    \includegraphics[width=\linewidth]{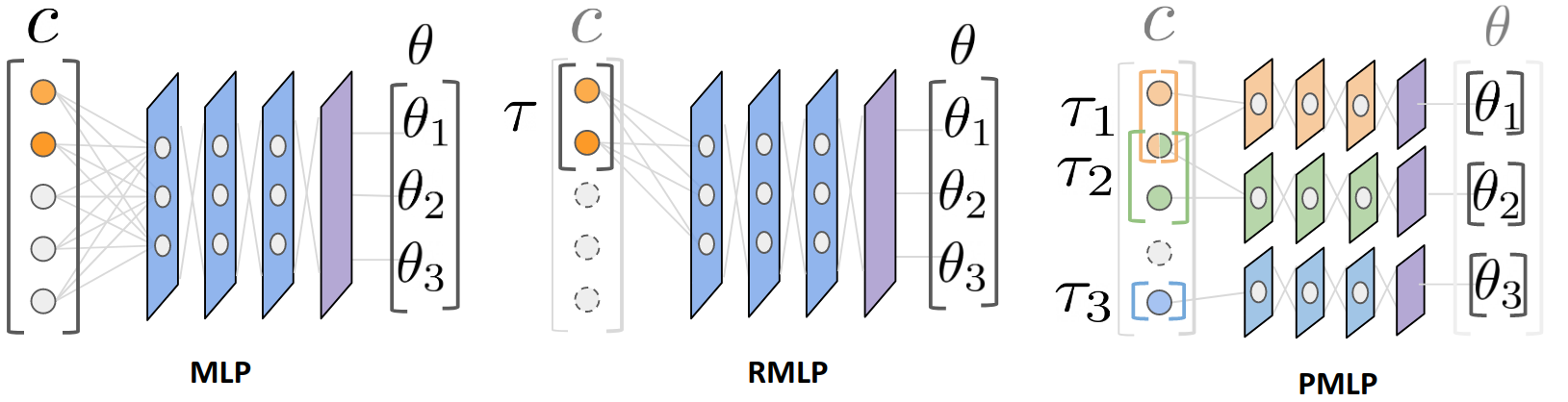}
    \caption{A visualization of the different policy types.
    \acronym is used to construct both the Reduced MLP (RMLP) and Partitioned MLP (PMLP). %
    The baseline MLP is also shown for comparison.
    The relevant states are also used for the critic portion of the networks (only the actor portion is shown).
    The notation used is [$w_1$,...,$w_d$], specifying the hidden units and depth of both the actor and critic.}
    \label{fig:networks}
    \vspace{-5mm}
\end{figure}

Given the full context $c$, the reduced context $\tau$, and the parameter-specific contexts $\tau_j$, the three MLP-based network architectures shown in Fig. \ref{fig:networks} are constructed and trained using actor-critic approaches~\cite{konda2000actor}.

The baseline $\pi(\theta|c)$ uses a standard MLP network.
The approach $\pi(\theta|\tau)$ uses a Reduced MLP (RMLP): an MLP network where the inputs are reduced to only the relevant context $\tau$.
The approach $\pi([\theta]_j|\tau_j)$ has independent sets of fully connected layers for each policy parameter $\theta_j$, but with potentially overlapping inputs depending on $\tau_j$. This Partitioned MLP (PMLP) network represents the structural causal model~\cite{pearl2009causality} for each $\theta_j$ with $\tau_j$ as parent variables.

To provide a fair comparison, we choose the weights for the PMLP according to heuristics and multiply the number of hidden units by the size of the action space to size the RMLP and MLP.
We originally sized the PMLP network according to~\cite{lu2017expressive} to provide theoretical guarantees regarding function approximation for one-element outputs (i.e., each $\theta_j$), but the resulting network size for the baseline MLP was intractable to train. %
All neural network weights are randomly initialized per orthogonal initialization~\cite{saxe2014exact} using $\sqrt{2}$ and $0.01$ for the scale terms of the hidden layers and output layers, respectively.
All networks use tanh activations.

\subsection{Network Training and Transfer}

For both the internal model and target domain, we train our policies using PPO~\cite{schulman2017proximal} with Stable Baselines~\cite{stable-baselines}. %
First, each network is pretrained with the internal model until the task family is solved.
Then, we transfer the network weights to the target domain and evaluate the policy to determine whether the policy transfers zero-shot.
Otherwise, fine-tuning is performed. %
Although freezing network layers has been explored for fine-tuning control policies~\cite{rusu2016progressive}, we permit the entire network to adapt because of the approximate nature of pretraining with the internal model.
The learned policy is considered to have solved the task family if it successfully achieves a predefined reward threshold on 50 validation tasks; this evaluation occurs after each policy update.

%% file: content/6_experiments.tex
\section{EXPERIMENTAL RESULTS}
\label{sec:experiments}

\begin{figure*}
    \centering
    \begin{subfigure}[b]{0.21\linewidth}
        \centering
        \includegraphics[height=1.2in]{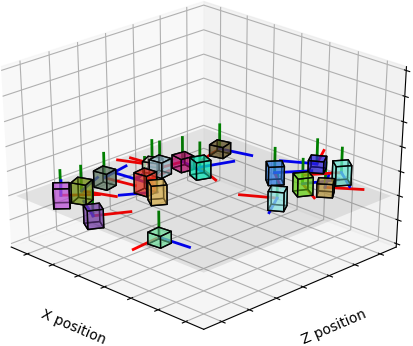}
        \caption{}
        \label{fig:blocks_im}
    \end{subfigure}
    \begin{subfigure}[b]{0.21\linewidth}
        \centering
        \includegraphics[height=1.2in]{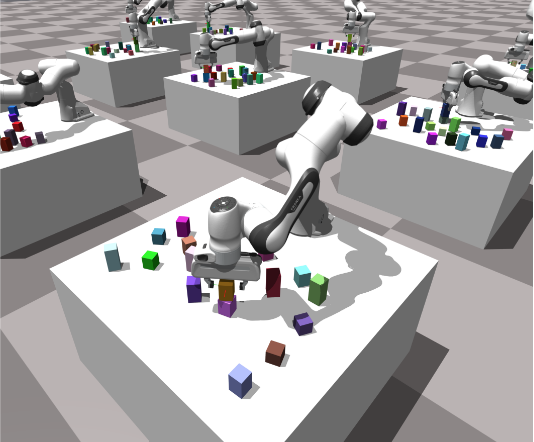}
        \caption{}
        \label{fig:blocks_isaac}
    \end{subfigure}
    \begin{subfigure}[b]{0.135\linewidth}
        \centering
        \includegraphics[height=1.2in]{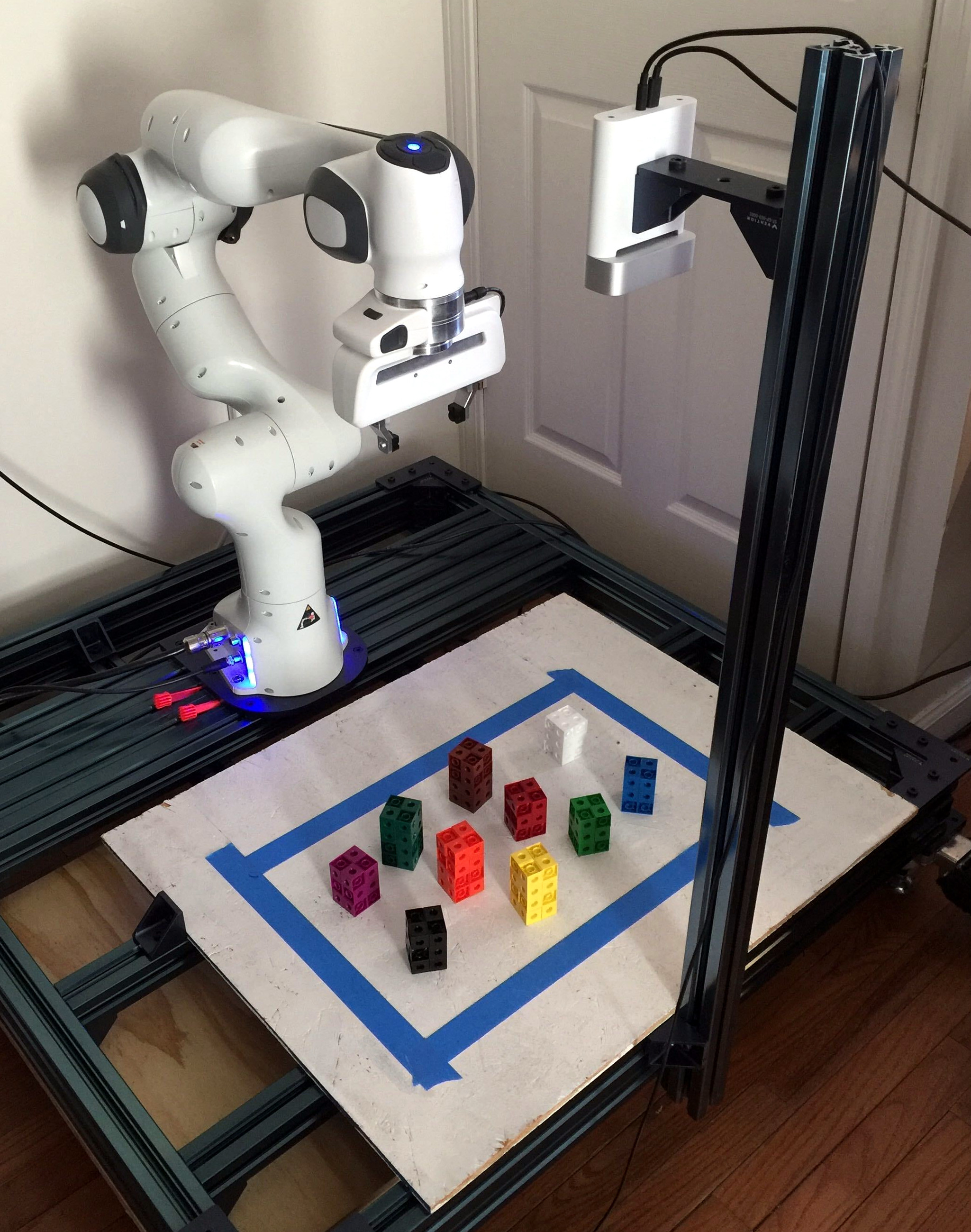}
        \caption{}
        \label{fig:franka_block_stack_exp}
    \end{subfigure}
    \begin{subfigure}[b]{0.21\linewidth}
        \centering
        \includegraphics[height=1.2in]{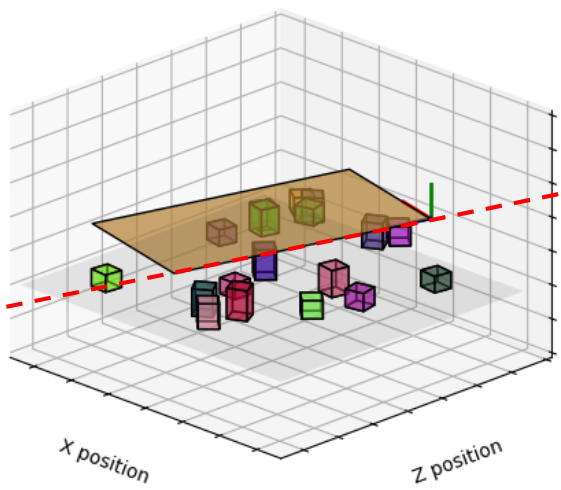}
        \caption{}
    \label{fig:crate_im}
    \end{subfigure}
    \begin{subfigure}[b]{0.21\linewidth}
        \centering
        \includegraphics[height=1.2in]{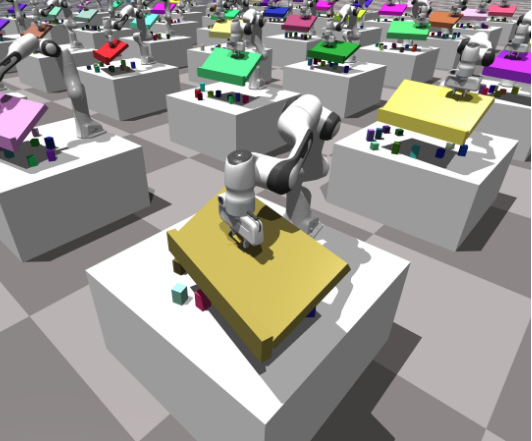}
        \caption{}
    \label{fig:crate_isaac}
    \end{subfigure}
    \caption{
    Transfer experiments for block stacking and crate opening manipulation tasks.
    Policies are pretrained in the internal model ((\subref{fig:blocks_im}),(\subref{fig:crate_im})) and then transferred to the target domain ((\subref{fig:blocks_isaac}),(\subref{fig:franka_block_stack_exp}),(\subref{fig:crate_isaac})).
    Target domains consist of replications of real systems using a Franka Panda robot, along with a real system for block stacking.
    Both tasks have distractor objects. For block stacking, only two blocks are necessary to generalize the policy. For crate opening, blocks represent distractor objects (e.g., if the crate were for a chest containing toys).
    }
    \label{fig:experiments} %
\end{figure*}

We evaluate how \acronym can construct policies with greater (target) sample-efficiency and robustness for the robot manipulation tasks of block stacking and crate opening (Fig.~\ref{fig:experiments}). %
Our experiments for each task follow the structure and transfer learning pipeline motivating our approach: 1) use \acronym to determine the causal structure of the task; 2) construct and pretrain policies with this structure; and 3) transfer and fine-tune these policies in the target domain. %
The target domains include manipulation tasks in NVIDIA Isaac Gym, enabling rigorous investigation of representative distribution shifts that may occur when deploying sim-to-real policies.
For the block stacking task, we additionally leverage a real robot system to assess sim-to-real transfer.

Target simulation and training was conducted using a NVIDIA DGX-1. Pretraining was done using a NVIDIA GeForce RTX 2080.
Samples from the block stacking and crate opening internal models were 400 and 65 times faster to obtain than target simulation samples, respectively.
This is consistent with the concept of the internal model as a cheap, approximate simulator, whereas samples from the target domain are costly and therefore desirable to minimize. %

Ten independent trials (from internal model pretraining to target fine-tuning) are conducted for each simulation experiment to provide statistically meaningful results given the variance inherent in model-free learning.
Statistics are provided in terms of mean and $\pm$1 standard deviation of policy updates requires to solve the task family. Samples are provided per 1000 (``k-Samples'') using a batch size of 512.

The supplementary materials describe further experiment details, such as the setup for the real block stacking target.

\subsection{Block Stacking}\label{sec:blocks}

The network architectures for the block stacking policies are specified in Table~\ref{tab:block_networks}. Although our policies are nonlinear, note this particular task is linear between $\tau$ and $\theta$.

\textbf{Task representation.} In the block stacking task, the context vector $c = [ \, c_{B_0}\transpose, \, \dots, \, c_{B_{N_B-1}}\transpose \, ] \transpose \in \mathbb{R}^{7 N_B}$ consists of the concatenation of $N_B$ individual block contexts.
The context vector for block $b$ is
$        c_{B_b} = [ \, 
        x_b^w, \,
        z_b^w, \,
        \psi_b, \,
        h_b, \,
        \bm{C}_b\transpose \, ] \transpose \in \mathbb{R}^{7}$
In the above equations, $x_b^w$ and $z_b^w$ are the world $x$- and $z$-positions of the blocks.
Each block orientation is defined by its rotation angle $\psi_b$ about the block's vertical axis ($y$).
The $y$-dimension, or height, of each block is $h_i$. %
Lastly, the block color $\bm{C}_b = [ \, R_b, \, G_b, \, B_b \, ]\transpose$ is specified via red-green-blue tuple.
Note that $y_b^w$ is not part of the context, as the initial scene always consists of blocks on the workspace plane.

The control policy $\pi(a|s,\theta_b)$ for block stacking is a sequential straight-line skill parameterized by
$\theta_{b} = [\, \theta_{\Delta x}, \, \theta_{\Delta y}, \, \theta_{\Delta z} \, ]\transpose \in \mathbb{R}^3$.
This skill specifies waypoints that the robot traverses via impedance control by lifting the source block vertically, moving horizontally, and descending to the desired location.
The skill preconditions are that the block is grasped and there are no obstructions to moving the object. 
The reward function is determined from the source block's position and the goal position upon the target block.

Using the internal model, \acronym correctly obtained the relevant context variables as  $\tau = [ x_0^w, x_1^w, h_1, z_0^w, z_1^w]^T$, $\tau_{\Delta x} = [ x_0^w, x_1^w]^T$, $\tau_{\Delta y} = [ h_1]$, and $\tau_{\Delta z} = [ z_0^w, z_1^w]^T$.

\begin{figure*}
    \centering
    \begin{subfigure}[b]{0.32\linewidth}
        \centering
        \includegraphics[width=0.95\linewidth]{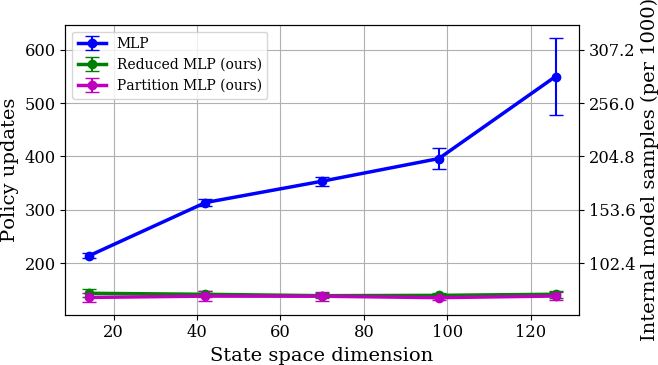}
        \caption{}
        \label{fig:block_scale_im_v0}
    \end{subfigure}
    \begin{subfigure}[b]{0.32\linewidth}
        \centering
        \includegraphics[width=0.95\linewidth]{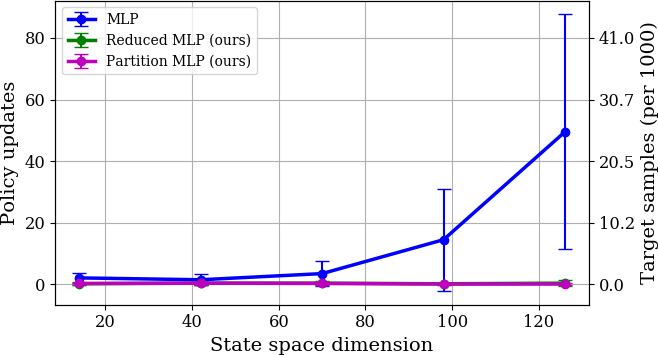}
        \caption{}
        \label{fig:block_scale_target_v0}
    \end{subfigure}
    \begin{subfigure}[b]{0.32\linewidth}
        \centering
        \includegraphics[width=0.99\linewidth]{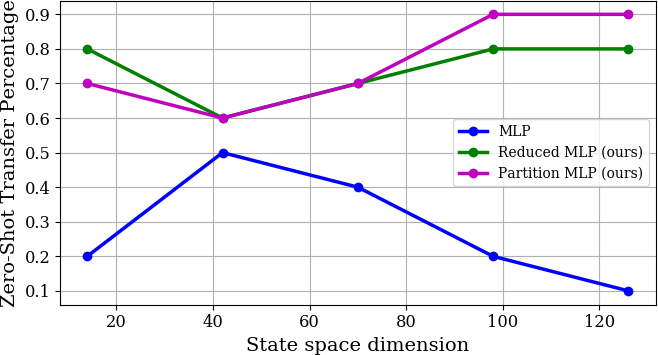}
        \caption{}
        \label{fig:block_scale_zeroshot_v0}
    \end{subfigure}
    \caption{Sample complexity of training a solved block stacking policy based on context dimension for (\subref{fig:block_scale_im_v0}) internal model and (\subref{fig:block_scale_target_v0}) target setting. \subref{fig:block_scale_zeroshot_v0}) Zero-shot transfer percentage, wherein the transferred policy needs no further target training to solve the task.}
    \label{fig:blocks_scale}
    \vspace{-5mm}
\end{figure*}

\textbf{Nominal transfer for increasing context size.} We conduct transfer experiments for $N_B = \{2, 6, 10, 14, 18\}$, with each $N_B$ conducted independently.
Our approach scales with the relevant part of the context space (Fig.~\ref{fig:blocks_scale}), bounding the sample requirements for the target (as well as the cheaper, internal model).
The increasing number of irrelevant dimensions from more blocks are eliminated by \acronym prior to conducting domain randomization during pretraining.

For the case of $N_B = 10$, we trained directly in the target domain without transfer and observed similar results as for pretraining, suggesting the internal model accords well with the target domain.
This explains why our approaches exhibit good zero-shot behavior over increasing context dimensions, unlike the baseline whose initial performance degrades as the number of irrelevant contexts increase.

\begin{table} %
    \centering
    \captionsetup{font=small}
    \caption{Networks used for the block stacking task.}
    \label{tab:block_networks}
    \begin{scriptsize}
    \begin{tabular}{c | c | c | c}
        Network & Parameters & Input Dim. (Total) & Architecture \\
        \hline
        MLP ($N_B=2$) & 3298 & 14 (14) & [24, 24, 24] \\
        MLP ($N_B=6$) & 4642 & 42 (42) & [24, 24, 24] \\
        MLP ($N_B=10$) & 5986 & 70 (70) & [24, 24, 24] \\
        MLP ($N_B=14$) & 7330 & 98 (98) & [24, 24, 24] \\
        MLP ($N_B=18$) & 8674 & 126 (126) & [24, 24, 24] \\
        RMLP (ours) & 2866 & 5 ($7 N_B$) & [24, 24, 24] \\
        PMLP (ours) & 754 & 5 ($7 N_B$) & [8, 8, 8] x 3 \\
    \end{tabular}
    \end{scriptsize}
    \vspace{-5mm}
\end{table}

\textbf{Distribution shift in irrelevant contexts.}
We now evaluate the robustness of the learned block stacking policy to distributions shifts in irrelevant context variables.
We conduct two transfer experiments, wherein the policies are pretrained using only half of the color space.
In the first case, the target has the same context distribution as the internal model.
In the second case, the target has the opposite color space (without overlap).
The experimental results (Table~\ref{tab:blocks_results_shift_color}) elucidate how a seemingly inconsequential variable can degrade policy execution through a distribution shift that the robot is not trained to expect.
Our approaches generate policies that are robust to these irrelevant domain shifts by construction; as \acronym explicitly identifies this dimension as unimportant and excludes it from target learning.

\begin{table*}
    \centering
    \captionsetup{font=small}
    \caption{Transfer results for a distribution shift in 30 context variables (color) that are irrelevant for the block stacking policy.}
    \label{tab:blocks_results_shift_color}
    \begin{scriptsize}
    \begin{tabular}{c || c || c | c || c | c }
        Network & \makecell{IM Updates \\ (k-Samples)} & \makecell{Target Updates \\ (k-Samples), no shift} & \makecell{Zero-Shot Transfer \\ no shift} & \makecell{Target Updates \\ (k-Samples), shift} & \makecell{Zero-Shot Transfer, \\ shift} \\
        \hline
        MLP & 237.5 (121.6) $\pm$ 11.6 (5.9) & 1.6 (0.8) $\pm$ 1.5 (0.8) & 4 & 17.3 (8.9) $\pm$ 4.8 (2.5) & 0\\
        RMLP (ours) & 137.6 (70.5) $\pm$ 7.2 (3.7) & 0.5 (0.3) $\pm$ 0.9 (0.5) & 7 & 1.0 (0.5) $\pm$ 1.5 (0.8) & 6\\
        PMLP (ours) & 139.2 (71.3) $\pm$ 8.2 (4.2) & 0.0 (0.0) $\pm$ 0.0 (0.0) & 10 & 0.1 (0.1) $\pm$ 0.3 (0.2) & 9\\
    \end{tabular}
    \end{scriptsize}
    \vspace{-3mm}
\end{table*}

\textbf{Sim-to-real policy evaluation.}\label{sec:real_blocks}
Lastly, to validate our approach for sim-to-real transfer, we evaluate the zero-shot policy performance on a real robot system that implements the block stacking task with $N_B = 10$ (Fig.~\ref{fig:franka_block_stack_exp}).
As shown in Table~\ref{tab:franka_block_stack_eval}, our policies successfully demonstrate greater zero-shot, sim-to-real transfer as compared to the baseline.

\begin{table} %
    \centering
    \captionsetup{font=small}
    \caption{
    Sim-to-real policy evaluation results for block stacking with $N_B = 10$.
    The reward threshold for zero-shot transfer is -0.025 (about half of the block width).
    We also note how often the block was successfully stacked.
    ``GT'' is a ground truth policy to illustrate the degree of uncertainty present within the robot perception and control system.
    Each policy was evaluated 10 times.} 
    \label{tab:franka_block_stack_eval}
    \begin{scriptsize}
    \begin{tabular}{c | c | c | c}
        Policy & Reward & Zero-Shot Transfer & Block Stacked \\ 
        \hline
        MLP & -0.033 $\pm$ 0.012 & 3 & 1 \\
        RMLP (ours) & -0.018 $\pm$ 0.007 & 9 & 4 \\
        PMLP (ours) & -0.014 $\pm$ 0.004 & 10 & 6 \\
        GT & -0.009 $\pm$ 0.003 & 10 & 10 \\
    \end{tabular}
    \end{scriptsize}
\end{table}

\subsection{Crate Opening}\label{sec:crate}

The crate opening experiment is nonlinear between $\tau$ and $\theta$, and the internal model, which is kinematic, presents a greater sim-to-real gap than block stacking.
Therefore, we also consider a second partitioned network, PMLP-R, with the same number of weights as the RMLP, to elicit possible influence of network expressivity in this domain due to the structural assumptions of the PMLP. 
The crate experiment primarily focuses on dynamics and context shifts, rather than varying numbers of objects, so unlike in blocks, the networks (Table~\ref{tab:crate_networks}) are the same for all experiments.
Beyond our two experiments in nominal transfer and dynamics shift, we also conducted a color shift experiment with similar results as with the blocks experiment.

\begin{table} %
    \centering
    \captionsetup{font=small}
    \caption{Networks used for the crate opening task.}
    \label{tab:crate_networks}
    \begin{scriptsize}
    \begin{tabular}{c | c | c | c}
        Network & Parameters & Input Dim. (Total) & Architecture \\
        \hline
        MLP & 13496 & 80 (80) & [40, 40, 40] \\
        RMLP (ours) & 7576 & 6 (80) & [40, 40, 40] \\
        PMLP (ours) & 1152 & 6 (80) & [8, 8, 8] x 5 \\
        PMLP-R (ours) & 8032 & 6 (80) & [24, 24, 24] x 5 \\
    \end{tabular}
    \end{scriptsize}
    \vspace{-5mm} %
\end{table}

\textbf{Task representation.} For the crate opening task, the context vector is $c = [ \, c_{C}\transpose, \, c_{B_0}\transpose, \dots, \, c_{B_{9}}\transpose \, ] \transpose \in \mathbb{R}^{80}$.
The block context is as defined previously. %
The crate context is $c_C= [ \, \bm{p}_C^w\transpose, \, \Phi, \, x_g^C, \, z_g^C, \, \Theta_o, \, \bm{C}_C\transpose \, ] \transpose \in \mathbb{R}^{10} $, 
where $\bm{p}_C^w = [ \, x_C^w, \, y_C^w, \, z_C^w \, ]\transpose$ is the position of the crate coordinate frame with respect to the world frame with vertical angle $\Phi$. The crate is always initially closed (horizontal), but the desired goal angle is specified by $\Theta_o$. The robot interacts with the crate via a grasp point specified in the frame of the crate by $x_g^C$ and $z_g^C$ which are orthogonal and parallel to the crate rotational axis, respectively. The color of the crate is $\bm{C}_C$. %

The control policy $\pi(a|s,\theta_a)$ is a robot skill that executes circular arcs emerging from the grasp point with the following parameterization:
$
    \theta_{a} = [\, \theta_{\bm{p}_a^w}\transpose, \, \theta_{\Delta \gamma}, \, \theta_{\Delta \phi} \, ]\transpose \in \mathbb{R}^5, 
    $
where $\theta_{\bm{p}_a^w} = [ \, \theta_{x_a^w}, \, \theta_{y_a^w}, \, \theta_{z_a^w} \, ]\transpose$ is the sphere position used to calculate the radius from the grasp point. Then, the arc is traced out $\theta_{\Delta \gamma}$ in azimuth and $\theta_{\Delta \phi}$ in inclination in polar coordinates from the grasp point. The skill preconditions are that the crate is grasped and unobstructed.
To learn this policy, the reward function is calculated from the crate angle error and total kinematic error. %

In this formulation, \acronym determined the following relevant context variables, which are expected based on rigid body articulation kinematics:
\begin{gather*}
    \tau = [ \, x_C^w, \, y_C^w, \, z_C^w, \, \Phi, \, z_g^C, \, \Theta_g ]^T \\
    \tau_{x_a^w} = [ x_C^w, \, \Phi, \, z_g^C ]^T,
    \tau_{y_a^w} = [ y_C^w ], 
    \tau_{z_a^w} = [ z_C^w, \, \Phi, \, z_g^C ]^T, \\
    \tau_{\Delta \gamma} = [ \Phi, \, \Theta_g  ]^T, 
    \tau_{\Delta \phi} = [ \Phi, \, \Theta_g  ]^T 
\end{gather*}

\textbf{Nominal transfer.} The transfer learning results for the crate opening policy is shown in Table~\ref{tab:crate_results_nominal_xfer}.
Pretraining the model reduced the number of target updates for the non-partitioned networks.
However, this was not the case for the partitioned networks, regardless of size.
This is likely a result of the discrepancy between the internal model and the target domain, which also explains the difference between the policy updates required for pretraining versus training directly in the target.
However, the reduction of relevant variables reduces the number of updates required to train directly in the target for both the RMLP and PMLP.

\begin{table}[t]
    \centering
    \captionsetup{font=small}
    \caption{Pretraining and transfer results for crate opening policies compared to training directly in target (without transfer).}
    \label{tab:crate_results_nominal_xfer}
    \begin{scriptsize}
    \begin{tabular}{c || c | c || c }
        Network &
            \makecell{IM Updates \\ (k-Samples)} &
            \makecell{Target Updates \\ (k-Samples), transfer} & 
            \makecell{Target Updates \\ (k-Samples), direct} \\
        \hline
        MLP  & 
            \makecell{45.0 $\pm$ 3.16 \\ (23.04 $\pm$ 1.62)} & 
            \makecell{12.20 $\pm$ 1.72 \\ (6.25 $\pm$ 0.88)} & 
            \makecell{38.70 $\pm$ 10.99 \\ (19.8 $\pm$ 5.63)} \\
        \hline
        \makecell{RMLP \\ (ours)} & 
            \makecell{32.40 $\pm$ 3.67 \\ (16.59 $\pm$ 1.88)} & 
            \makecell{7.0 $\pm$ 1.18 \\ (3.58 $\pm$ 0.61)} & 
            \makecell{16.40 $\pm$ 4.05 \\ (8.40 $\pm$ 2.08)} \\
        \hline
        \makecell{PMLP \\ (ours)} & 
            \makecell{48.20 $\pm$ 13.33 \\ (24.68 $\pm$ 6.83)} & 
            \makecell{14.0 $\pm$ 3.74 \\ (7.17 $\pm$ 1.92)} & 
            \makecell{14.20 $\pm$ 6.32 \\ (7.27 $\pm$ 3.24)} \\
        \hline
        \makecell{PMLP-R \\ (ours)} & 
            \makecell{51.30 $\pm$ 10.82 \\ (26.27 $\pm$ 5.54)} &
            \makecell{14.3 $\pm$ 4.34 \\ (7.32 $\pm$ 2.22)} & 
            \makecell{15.80 $\pm$ 3.97 \\ (8.09 $\pm$ 2.03)} \\
    \end{tabular}
    \end{scriptsize} %
\end{table}

\textbf{Dynamics distribution shift.}
Unlike the block stacking problem, the modeling gap between the internal model and target setting is sufficiently large that the trained policies incur a significant performance degradation upon first evaluating in the target domain.
We investigated this further by transferring the policies to two target settings with different crate stiffness values.
To focus on this dynamics shift, no other shifts (e.g., in context space) were induced. %

\begin{table}
    \centering
    \captionsetup{font=small}
    \caption{Fine-tuning for crate opening policies with increasing crate stiffness and correspondingly greater transition model difference between the internal model and target task.}
    \label{tab:crate_results_stiffness}
    \begin{scriptsize}
    \begin{tabular}{c || c || c || c }
        Network & 
            \makecell{Target Updates \\ (k-Samples), light} & 
            \makecell{Target Updates \\ (k-Samples), nominal} & 
            \makecell{Target Updates \\ (k-Samples), stiff} \\
        \hline
        MLP  & 
            \makecell{3.90 $\pm$ 0.54 \\ (2.00 $\pm$ 0.28)} &
            \makecell{12.20 $\pm$ 1.72 \\ (6.25 $\pm$ 0.88)} &
            \makecell{27.30 $\pm$ 5.51 \\ (13.98 $\pm$ 2.82)} \\
        \hline
        \makecell{RMLP \\ (ours)} & 
            \makecell{3.20 $\pm$ 0.60 \\ (1.64 $\pm$ 0.31)} &
            \makecell{7.00 $\pm$ 1.18 \\ (3.58 $\pm$ 0.61)} &
            \makecell{16.40 $\pm$ 5.90 \\ (8.40 $\pm$ 3.02)} \\
        \hline
        \makecell{PMLP \\ (ours)} & 
            \makecell{9.10 $\pm$ 1.58 \\ (4.66 $\pm$ 0.81)} &
            \makecell{14.00 $\pm$ 3.74 \\ (7.17 $\pm$ 1.92)} &
            \makecell{24.00 $\pm$ 15.06 \\ (12.29 $\pm$ 7.71)} \\
        \hline
        \makecell{PMLP-R \\ (ours)} & 
            \makecell{9.20 $\pm$ 1.54 \\ (4.71 $\pm$ 0.79)} &
            \makecell{14.30 $\pm$ 4.34 \\ (7.32 $\pm$ 2.22)} &
            \makecell{19.10 $\pm$ 4.91 \\ (9.80 $\pm$ 2.51)} \\
    \end{tabular}
    \end{scriptsize}
    \vspace{-3mm}
\end{table}

The results in Table~\ref{tab:crate_results_stiffness} suggest that increasing the stiffness is sufficient as a proxy for increasing the modeling difference between the internal model and the target.
The optimal parameters for the kinematic case (internal model) are not necessarily the same as the target domain with realistic dynamics of manipulation using impedance control.
Therefore, greater modeling differences implies that greater search in policy parameter space is required to converge to parameters that generalize in the target domain.

In all cases, we see that the RMLP network performs best.
As a likely consequence of a less expressive network with a larger dynamics gap, the smaller PMLP network demonstrated a significant variance increase in the higher stiffness case than the larger PMLP-R. %
Overall, our policies are more robust to distribution shifts in model dynamics.
However, we note that partitioning imposes structure that may not be optimal for this problem, as the MLP outperformed the partitioned networks in the light and nominal stiffness cases. %

%% file: content/7_conclusion.tex
\section{CONCLUSION}\label{sec:conclusion}
The causal reasoning afforded by \acronym allows the robot to structure robot manipulation policies with fewer parameters that are more sample efficient and robust to domain shifts than a naive approach that includes all known contexts. %
Indeed, using causality to reason about the simulation of a task identifies \textit{what} variables are important to generalize a policy, while domain randomized pretraining provides a strong, task-specific prior in terms of \textit{how} they matter.
We believe that \acronym is one step towards a new paradigm for structural sim-to-real transfer of robot manipulation policies that are sufficiently lightweight to be adapted in-the-field to overcome unforeseen domain shifts.

For future work, we will investigate using precondition learning to relax the assumption that the policy execution is feasible.
We will also explore how the robot can learn the internal model used as the causal reasoning engine. %

%% file: content/8_acknowledgments.tex
\section*{ACKNOWLEDGMENTS}
We gratefully acknowledge support from the U.S.~Office of Naval Research (Grant N00014-18-1-2775), U.S.~Army Research Laboratory (Grant W911NF-18-2-0218 as part of the A2I2 Program), and the NVIDIA NVAIL Program.

%% file: content/9_appendix.tex
\begin{appendices}

\section*{Supplementary Materials}

\subsection{Summary of CREST}

Which state features are important for learning a control policy?
Our approach, CREST, addresses this question through causal feature selection.
CREST selects the relevant state variables for a given control policy, which apply over the policy's preconditions.
The assumptions for CREST are that an internal model (i.e., an approximate task simulation) exists, the context space representation of the internal model facilitates causal interventions (e.g., disentangled variables), and the (parameterized) control policy and its preconditions are known.
Through structure and transfer learning, CREST enables learning of policies that are compact, avoiding unnecessary state features.
By construction, policies built using CREST are robust to distribution shifts in irrelevant variables, whereas baseline methods may yield policies with spurious correlations that are brittle.
Such distribution shifts could arise from transfer between the internal model and reality, due to variations in dynamics or context distributions not encountered during pretraining with the internal model.

\subsection{CREST Analysis on Math Environment}\label{sec:matrix_env_exp}

We now provide a greater description of the manipulation environment described in Sec.~\ref{sec:crest_eval}.
The toy environment, \texttt{MathManipEnv}, approximates the mathematics of a controller for goal-based manipulation.
For simplicity, the low-level control policy simply perturbs the state $s \in \mathbb{R}^{|S|}$ by an input of $\theta = a \in \mathbb{R}^{|A|}$ in a manner specific to whether the system is linear or non-linear.
Additionally, we consider the context $c \in \mathbb{R}^{|S|}$ to be the initial state, $s_0$. For this evaluation, we considered cases where $|S| = |A|$ (``Dim.'' in Table~\ref{tab:causal_analysis}).

The reward for this task is
\begin{align*}
    r &= - \| \bm{g}_a - \bm{g}_d \| \\
      &= - \| G s_a - \bm{g}_d \| \\
      &= - \| G (s_0 + A(\theta)) - \bm{g}_d \|
\end{align*}
where $\bm{g}_a \in \mathbb{R}^{|g|}$ is the goal vector that was obtained after execution of the controller to yield achieved state $s_a$, and $\bm{g}_d \in \mathbb{R}^{|g|}$ is the desired goal.
The goal vector is calculated from a \textit{goal selection matrix} $G \in \mathbb{R}^{|g| \times |S|}$, which is a one-hot encoding matrix where the columns indicate the elements of the state vector that are used.
In practice, $G$ is formed by first randomly selecting $N_\tau$ relevant context variables from the total set of $c$ to form $\tau$.
Then, each $\tau$ is randomly allocated to a separate dimension of the goal vector, i.e., row of $G$.
Here, $G$ represents that, in some goal-based problems, the goal is calculated from only a subset of the state vector (e.g., relative to the position of a particular object).

The process of the system is either linear or non-linear, where $A(\theta) = \Delta s + w_a$ and $w_a \sim \mathcal{N}(0, \sigma_a^2)$ is the action noise.
In the linear case, the controller $A \in \mathbb{R}^{|S| \times |S|}$ is a matrix with randomly selected coefficients, so $\Delta s = A \theta$.
The non-zero coefficients of $A$ indicate mappings of $\tau_j$ to $\theta_j$.
In the non-linear case, the controller $A$ is a list of size $|A|$, where each element of the list specifies randomly selected functions (exponential, sigmoid, sine, cosine) that transform input $a_j$ into the resulting $\Delta s_j$.

Each trial of this environment randomly selects different $\tau$, $\bm{g}_d$, $s_0$, $G$, and $A$.
The goal vector dimensionality $|g|$ is fixed for each run and is typically equal to $|S|$.

\subsection{Task Representation: Block Stacking}\label{sec:task_rep_block}

\begin{figure}[b]
    \centering
    \includegraphics[width=\linewidth]{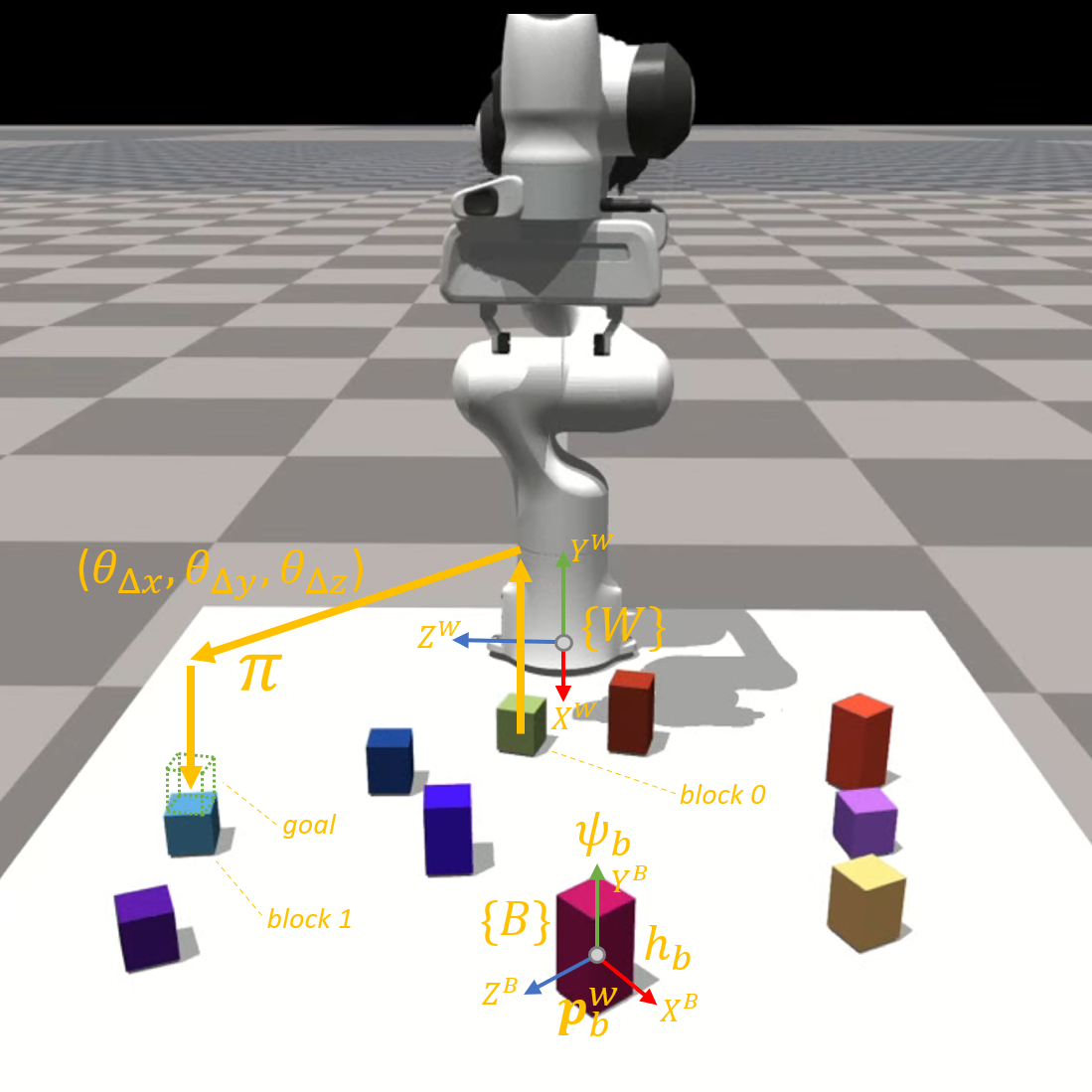}
    \caption{Diagram of the block stacking task. The world coordinate frame $\{W\}$ is defined at the base link of the robot. Each block coordinate frame $\{B\}$ is defined at the block's centroid.}
    \label{fig:blocks_diagram}
\end{figure}

We now provide additional detail for the block stacking task described in Sec.~\ref{sec:blocks}.
Figure~\ref{fig:blocks_diagram} illustrates some context variables and the policy trajectory.
In this task, the robot must stack the source block (block 0) upon the target block (block 1) using a sequential straight-line skill with control policy $\pi(a|s,\theta_{b})$ and known preconditions.

\textbf{Policy.} The policy parameters $\theta_{b} = [\, \theta_{\Delta x}, \, \theta_{\Delta y}, \, \theta_{\Delta z} \, ]\transpose \in \mathbb{R}^3$ define three waypoints that the robot is sequentially commanded to via impedance control. Specifically, let $y_p$ represent a vertical position above the table and blocks. After the block is grasped, the executed policy is therefore:
\begin{enumerate}
    \item Vertically lift to $y_p$.
    \item Move $(\theta_{\Delta x}, \theta_{\Delta z})$ at fixed $y_p$.
    \item Vertically move $\theta_{\Delta y} - y_p$.
\end{enumerate}
The vertical lift to $y_p$ avoids obstructions to moving the block, so the preconditions of this skill are always satisfied.

\textbf{Reward.} The reward function for this task is
\begin{gather*}
    r = - \alpha \| \bm{p}_{0,a}^w - \bm{p}_g^w \|
\end{gather*}
where $\alpha = 1$ is the reward weight, $\bm{p}_g^w = [ \, x_g^w, \, y_g^w, \, z_g^w \, ]\transpose$ is the goal position of block 0, and $\bm{p}_{0,a}^w = [ \, x_{0,a}^w, \, y_{0,a}^w, \, z_{0,a}^w \, ]\transpose$ is the final (i.e., achieved) position of block 0 at the end of the policy execution.
In this task, the goal is to stack block 0 upon block 1.
Therefore, $x_g^w = x_1^w$, 
$y_g^w = \tfrac{1}{2} h_0 + \tfrac{1}{2} h_1 + y_1^w$, and
$z_g^w = z_1^w$.

Expressing the reward function in terms of the task context variables elucidates which variables are considered relevant to the task policy. The optimal low-level policy parameters $\theta_{b}^* = [\, \theta_{\Delta x}^*, \, \theta_{\Delta y}^*, \, \theta_{\Delta z}^* \, ]\transpose \in \mathbb{R}^3$ are
\begin{align*}
    \theta_{\Delta x}^* &= x_g^w - x_0^w \\
                        &= x_1^w - x_0^w \\
    \theta_{\Delta y}^* &= y_g^w - y_0^w \\
                        &= \tfrac{1}{2} h_0 + \tfrac{1}{2} h_1 + y_1^w - y_0^w \\
                        &= \tfrac{1}{2} h_0 + \tfrac{1}{2} h_1 + \tfrac{1}{2} h_1 - \tfrac{1}{2} h_0 \\
                        &= h_1 \\
    \theta_{\Delta z}^* &= z_g^w - z_0^w \\
                        &= z_1^w - z_0^w
\end{align*}
where the reduction of the block $y$-position variables arise from the blocks being initially constrained to the table.

The above derivation demonstrates that only certain variables are needed to generalize the policy across different contexts where the preconditions also hold true.
Moreover, certain variables are only influential in certain policy parameters.
We express this more concretely by formalizing what variables are needed for each parameter, which is where the ground truth mappings for CREST (Sec.~\ref{sec:blocks}) arise:
\begin{align*}
    \theta_{\Delta x}^* = f(x_0^w, x_1^w) &\rightarrow \tau_{\Delta x}^* = [ x_0^w, x_1^w ]^T \\
    \theta_{\Delta y}^* = f(h_1) &\rightarrow \tau_{\Delta y}^* = [ h_1 ] \\
    \theta_{\Delta z}^* = f(z_0^w, z_1^w) &\rightarrow \tau_{\Delta z}^* = [ z_0^w, z_1^w ]^T
\end{align*}
\begin{align*}
    \theta_{b}^* = f(x_0^w, x_1^w, h_1, z_0^w, z_1^w) \rightarrow \tau^* = [ x_0^w, x_1^w, h_1, z_0^w, z_1^w]^T
\end{align*}

Here, $f$ is the model, which for our work we characterize using a neural network (although for this specific task, a linear model would also suffice). In causality terms, this is equivalent to modeling each individual policy parameter ($\theta_j$) as a structural causal model, where $f$ is a function with parent variables given by $\tau_j$.

\subsection{Sim-to-Real Block Stacking Experiment}\label{sec:sim_to_real_exp}

The sim-to-real block stacking experiment (Sec.~\ref{sec:real_blocks}) demonstrates that our proposed approach works in practice on a real robot system (Fig.~\ref{fig:franka_block_stack_exp}).
As it is experimentally difficult to realize all possible values within the context distributions (e.g., creating blocks of precise height and color for each sample), we instead conduct the experiment on a slightly reduced distribution range.
Specifically, we conduct this experiment using 10 blocks, where each block has a different color and two possible heights (5.7 cm or 7.6 cm).
Before each trial, all block positions and rotations are shuffled by hand.
Additionally, a random number generator selects the height of each block, as well as the enumeration of the blocks (and therefore which blocks are the source and target).
The length and width of each block is 4.2 cm, which does not change during the experiment and is known from manual measurement (i.e., not perception).

\begin{figure}
    \centering
    \includegraphics[width=\linewidth]{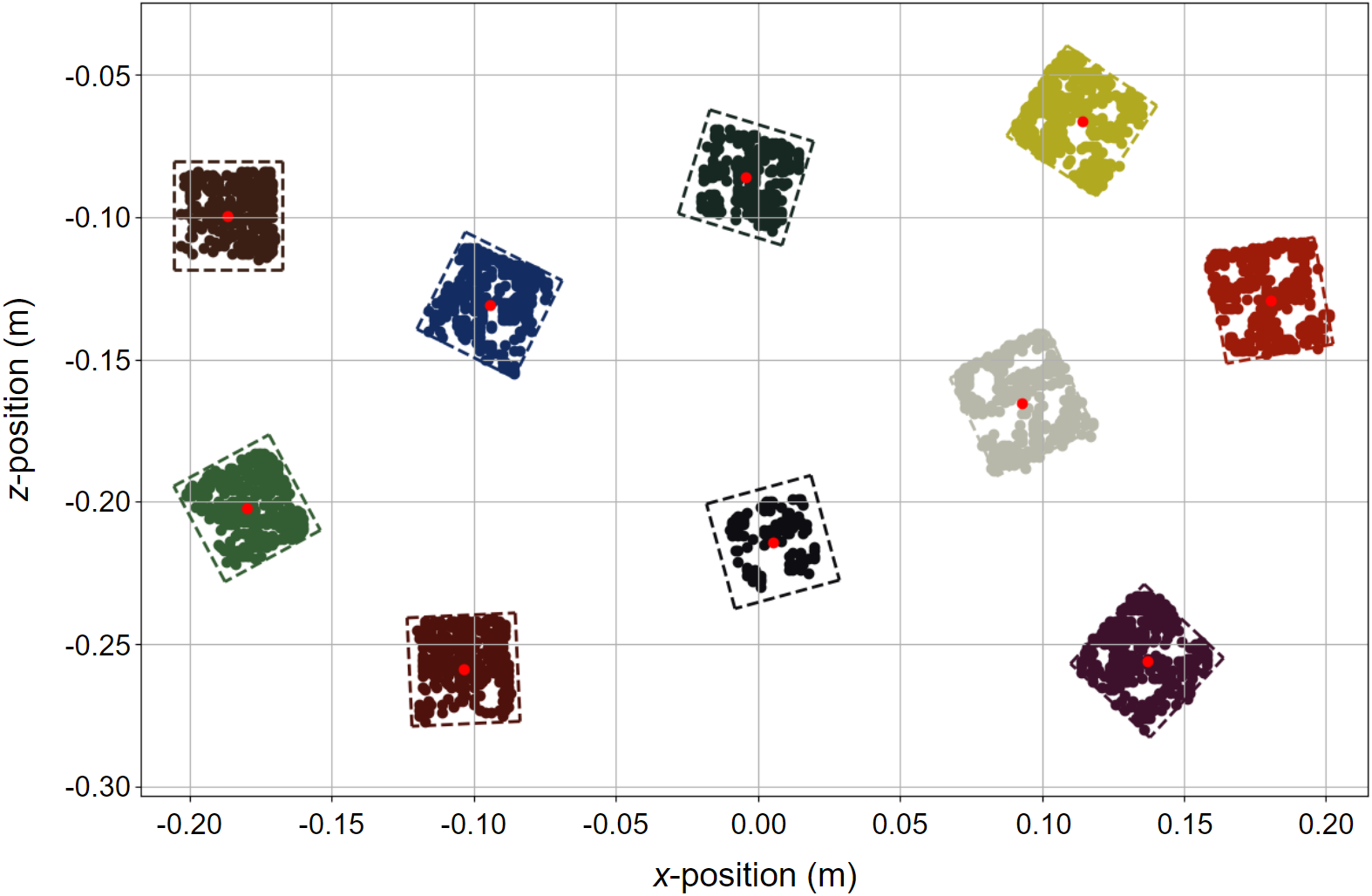}
    \caption{
        Block state estimation used for the sim-to-real experiments using RGB-D perception. %
        The perception algorithm takes as input a colored point cloud, and outputs a position, rotation, and color for 10 blocks.
        The red point in each block represents the block centroid, and the dashed lines indicate the block length and width (known \textit{a priori}).
        The best-fit position and rotation angle for each point cloud cluster yields a pose estimate for each block.
    }
    \label{fig:blocks_perception}
\end{figure}

\textbf{Perception.}
We use a Microsoft Azure Kinect RGB-D camera to estimate each block's position, rotation, and color through a model-based perception algorithm utilizing the Open3D library~\cite{Zhou2018}.
Figure~\ref{fig:blocks_perception} shows an example of the block perception.
The perception algorithm is as follows:
\begin{enumerate}
    \item Crop to region bounded by the table blue tape (Fig.~\ref{fig:franka_block_stack_exp}).
    \item Removal of hidden points via Katz~\cite{katz2007direct}, i.e., points expected to be occluded from the camera viewpoint.
    \item Fit plane to table via random sample consensus (RANSAC) and remove any points below this plane.
    \item Detect remaining clusters with DBSCAN~\cite{ester1996density}, a density-based clustering algorithm. Proceed only if the numbers of clusters is $N_B = 10$, or reject the perception sample and try again.
    \item For each cluster (block), determine the best position and angle that fits a cube of known dimensions to the cluster via least-squares optimization. This step yields an estimate of each block's position and rotation.
    \item Estimate block color by averaging the colors of all points within a cluster (block).
\end{enumerate}
Due to difficulties with accurately estimating block height from depth, the block height is provided by manual input instead. Manual checks are also completed prior to executing the control policy to ensure block perception results are reasonable. For example, if one cluster was not a block, but part of the blue tape, the perception sample would be rejected and attempted again. Prior to running the perception system, we obtain the extrinsics of the camera via a target-based calibration procedure, and we use the intrinsics as reported directly from the camera.

\textbf{Control.}
We use the \texttt{FrankaPy} library~\cite{zhang2020modular} that implements impedance control for the Franka Emika Panda robot.

\subsection{Task Representation: Crate Opening}\label{sec:task_rep_crate}

\begin{figure}%
    \centering
    \includegraphics[width=\linewidth]{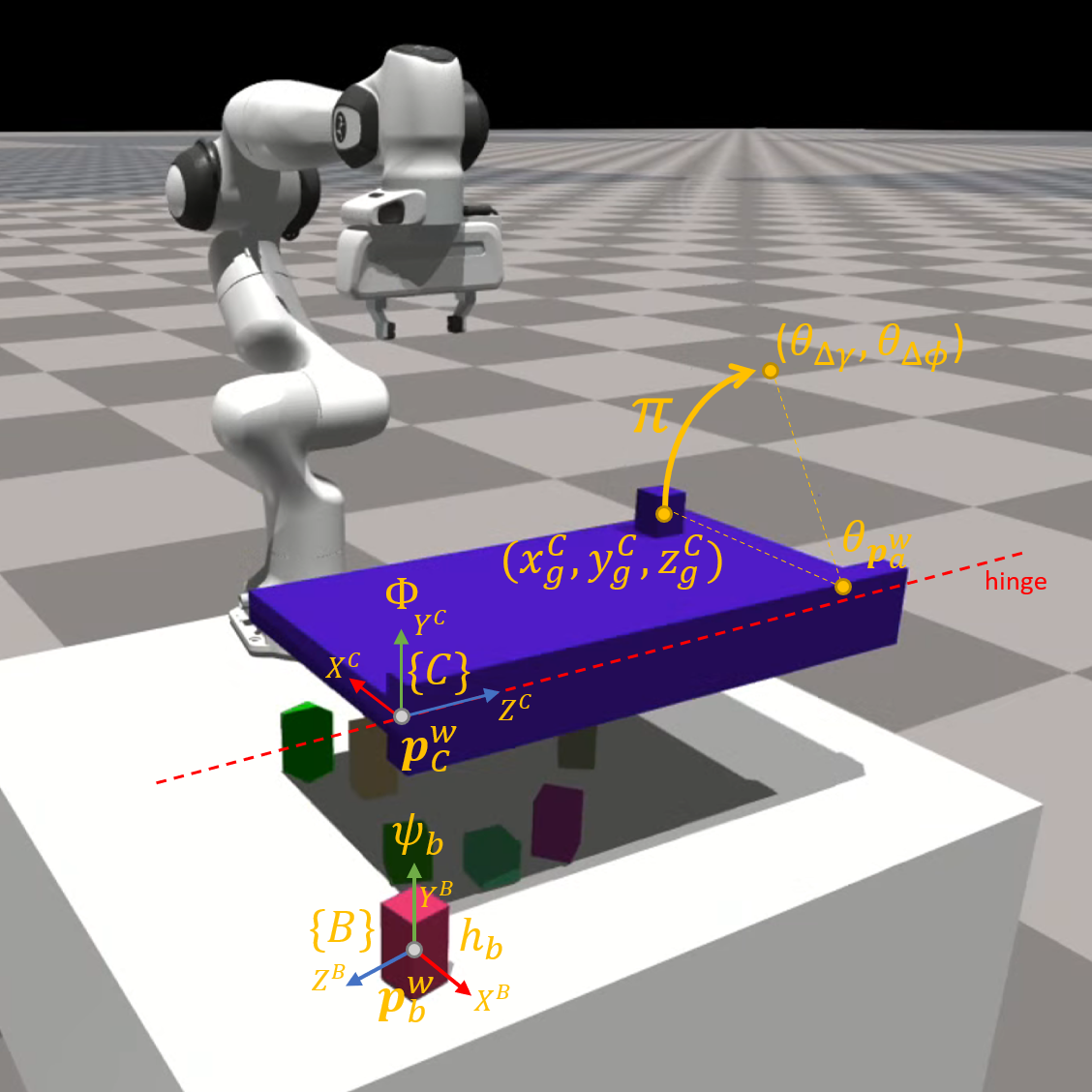}
    \caption{Diagram of the crate opening task. The world coordinate frame $\{W\}$ (not shown) is defined at the base link of the robot, similar to the block stacking task (Fig.~\ref{fig:blocks_diagram}). The $z$-axis of the crate coordinate frame $\{C\}$ is coincident with the crate hinge. There are 10 distractor blocks, each with coordinate frame $\{B\}$.}
    \label{fig:crate_diagram}
\end{figure}

This section provides more detail of the crate opening task (Sec.~\ref{sec:crate}), where the objective is to open a crate in the presence of distractor objects using a circular arc skill with control policy $\pi(a|s,\theta_a)$ and known preconditions.
Figure~\ref{fig:crate_diagram} shows some context variables and the policy trajectory, which emerges from the crate grasp point. %

This task has a larger modeling difference between the internal model and the target domain, which could also contribute to why our partitioned networks (PMLP, PMLP-R) were less successful than our non-partitioned network (RMLP).
In addition to the dynamics domain difference discussed in Sec.~\ref{sec:crate}, the $y$-position of the grasp point, $y_g^C$, is also slightly different. For the internal model, $y_g^C$ exists in the same plane as the crate, but for the target, $y_g^C$ is slightly above the crate because of the protruding grasp point.

\textbf{Policy.} The policy parameters $\theta_{a} = [\, \theta_{\bm{p}_a^w}\transpose, \, \theta_{\Delta \gamma}, \, \theta_{\Delta \phi} \, ]\transpose \in \mathbb{R}^5$ define a circular arc that is composed of $N_T$ waypoints.
The robot is commanded to the crate grasp point, then the robot executes the policy by sequentially following each waypoint via impedance control.
The crate cannot open into the blocks below, so the preconditions are always satisfied.

\textbf{Reward.} The reward function for this task is
\begin{gather*}
    r = - \| [ \, \alpha_a \Delta \Theta , \, \alpha_k e_k \, ]\transpose \|
\end{gather*}
In this function, $\Delta \Theta = \Theta_a - \Theta_o $ is the difference between the achieved ($\Theta_a$) and goal ($\Theta_o$) crate angles, and $\alpha_a$ and $\alpha_k$ are reward weights.
The term $e_k$ is the kinematic error in the policy trajectory, which is intended to induce robot trajectories that are safe (physically realizable and low force) in the target domain given articulated motion of the crate.
For this work, $\alpha_a = 1$ and $\alpha_k$ is 5 for the internal model and 0 for the target domain (because the robot realizes the trajectory it can actually achieve on the target due to the crate's articulated motion). %

Specifically, $e_k = \tfrac{1}{N_T}\sum_{t = 1}^{N_T} \| \bm{p}_{a,t}^w - \bm{p}_{d,t}^w \|$, where $\bm{p}_{d,t}^w$ is the desired position of a waypoint in the trajectory and $\bm{p}_{a,t}^w$ is the kinematically realizable position of that same waypoint, both at timestep $t$.
This is determined by projecting the desired waypoint onto the plane formed by rotating the grasp point about the crate hinge, obtaining the resulting crate angle, and using this angle to compute the realized grasp point.

\subsection{Crate Opening Distribution Shift in Irrelevant Contexts}
As mentioned in Sec.~\ref{sec:crate}, we also conducted a crate opening experiment with distribution shifts in irrelevant parts of the context space, similar to the experiment in the block stacking task (Table~\ref{tab:blocks_results_shift_color}).
As before, we pretrain on the entire context space, except for color of the crate and blocks, where only half of the color space is used.
For testing, we transfer to two cases: 1) the same color space seen in training (no shift), and 2) the opposite color space (complete shift with no overlap).
This experiment uses the ``light'' crate stiffness.

Table~\ref{tab:crate_results_shift_color} shows the results of this experiment.
As expected, our policies are robust to distribution shifts of this type, whereas the baseline MLP incurs approximately 55\% more target updates to overcome these irrelevant distribution shifts. %
Unlike the version of this experiment for block stacking, no policies achieved zero-shot transfer.
However, this is because of the previously described domain shift in dynamics between the internal model and target domain. %

\begin{table}%
    \centering
    \captionsetup{font=small}
    \caption{Transfer results for a distribution shift in 33 context variables that are irrelevant for the crate opening policy. Variables are 3-tuple RGB colors of the crate and 10 blocks in the scene. Light crate stiffness.}
    \label{tab:crate_results_shift_color}
    \begin{scriptsize}
    \begin{tabular}{c || c || c || c}
            Network & 
            \makecell{IM Updates \\ (k-Samples)} &
            \makecell{Target Updates \\ (k-Samples), no shift} &
            \makecell{Target Updates \\  (k-Samples), shift} \\
        \hline
            MLP & 
            \makecell{36.10 $\pm$ 3.11 \\ (18.48 $\pm$ 1.59)} &
            \makecell{11.10 $\pm$ 2.30 \\ (5.68 $\pm$ 1.18)} &
            \makecell{17.30 $\pm$ 4.22 \\ (8.86 $\pm$ 2.16)} \\
        \hline
            \makecell{RMLP \\ (ours)} & 
            \makecell{36.20 $\pm$ 5.23 \\ (18.53 $\pm$ 2.68)} & 
            \makecell{3.40 $\pm$ 0.66 \\ (1.74 $\pm$ 0.34)} & 
            \makecell{3.50 $\pm$ 0.67 \\ (1.80 $\pm$ 0.34)} \\
        \hline
            \makecell{PMLP \\ (ours)} & 
            \makecell{48.50 $\pm$ 7.88 \\ (24.80 $\pm$ 4.03)} & 
            \makecell{8.50 $\pm$ 2.06 \\ (4.35 $\pm$ 1.06)} & 
            \makecell{8.30 $\pm$ 1.10 \\ (4.25 $\pm$ 0.56)} \\
        \hline
            \makecell{PMLP-R \\ (ours)} & 
            \makecell{53.00 $\pm$ 7.80 \\ (27.14 $\pm$ 3.99)} & 
            \makecell{9.50 $\pm$ 1.91 \\ (4.86 $\pm$ 0.98)} & 
            \makecell{9.60 $\pm$ 2.01 \\ (4.92 $\pm$ 1.03)} \\
    \end{tabular}
    \end{scriptsize}
\end{table}

\end{appendices}

%% file: main.bbl
\begin{thebibliography}{10}
\providecommand{\url}[1]{#1}
\csname url@rmstyle\endcsname
\providecommand{\newblock}{\relax}
\providecommand{\bibinfo}[2]{#2}
\providecommand\BIBentrySTDinterwordspacing{\spaceskip=0pt\relax}
\providecommand\BIBentryALTinterwordstretchfactor{4}
\providecommand\BIBentryALTinterwordspacing{\spaceskip=\fontdimen2\font plus
\BIBentryALTinterwordstretchfactor\fontdimen3\font minus
  \fontdimen4\font\relax}
\providecommand\BIBforeignlanguage[2]{{%
\expandafter\ifx\csname l@#1\endcsname\relax
\typeout{** WARNING: IEEEtran.bst: No hyphenation pattern has been}%
\typeout{** loaded for the language `#1'. Using the pattern for}%
\typeout{** the default language instead.}%
\else
\language=\csname l@#1\endcsname
\fi
#2}}

\bibitem{peng2018sim}
X.~B. Peng, M.~Andrychowicz, W.~Zaremba, and P.~Abbeel, ``{Sim-to-Real Transfer
  of Robotic Control with Dynamics Randomization},'' \emph{Int'l Conf. on
  Robotics and Automation (ICRA)}, 2018.

\bibitem{bousmalis2018using}
K.~Bousmalis, A.~Irpan, P.~Wohlhart, Y.~Bai, M.~Kelcey, M.~Kalakrishnan,
  L.~Downs, J.~Ibarz, P.~Pastor, K.~Konolige, \emph{et~al.}, ``{Using
  Simulation and Domain Adaptation to Improve Efficiency of Deep Robotic
  Grasping},'' \emph{Int'l Conf. on Robotics and Automation (ICRA)}, 2018.

\bibitem{kroemer2019review}
O.~Kroemer, S.~Niekum, and G.~Konidaris, ``{A Review of Robot Learning for
  Manipulation: Challenges, Representations, and Algorithms},'' \emph{arXiv
  preprint arXiv:1907.03146}, 2019.

\bibitem{zhao2020sim}
W.~Zhao, J.~P. Queralta, and T.~Westerlund, ``{Sim-to-Real Transfer in Deep
  Reinforcement Learning for Robotics: a Survey},'' \emph{IEEE Symposium Series
  on Computational Intelligence (SSCI)}, 2020.

\bibitem{tobin2017domain}
J.~Tobin, R.~Fong, A.~Ray, J.~Schneider, W.~Zaremba, and P.~Abbeel, ``{Domain
  Randomization for Transferring Deep Neural Networks from Simulation to the
  Real World},'' \emph{Int'l Conf. on Intelligent Robots and Systems (IROS)},
  2017.

\bibitem{liang2018gpu}
J.~Liang, V.~Makoviychuk, A.~Handa, N.~Chentanez, M.~Macklin, and D.~Fox,
  ``{GPU-Accelerated Robotic Simulation for Distributed Reinforcement
  Learning},'' \emph{Conf. on Robot Learning (CoRL)}, 2018.

\bibitem{zhang2018learning}
K.~Zhang, B.~Sch{\"o}lkopf, P.~Spirtes, and C.~Glymour, ``{Learning Causality
  and Causality-Related Learning: Some Recent Progress},'' \emph{{National
  Science Review}}, vol.~5, no.~1, pp. 26--29, 2018.

\bibitem{scholkopf2019causality}
B.~Sch{\"o}lkopf, ``{Causality for Machine Learning},'' \emph{arXiv preprint
  arXiv:1911.10500}, 2019.

\bibitem{zhang2017transfer}
J.~Zhang and E.~Bareinboim, ``{Transfer Learning in Multi-Armed Bandits: A
  Causal Approach},'' \emph{Int'l Joint Conf. on Artificial Intelligence
  (IJCAI)}, 2017.

\bibitem{de2019causal}
P.~de~Haan, D.~Jayaraman, and S.~Levine, ``{Causal Confusion in Imitation
  Learning},'' \emph{Conf. on Neural Information Processing Systems (NeurIPS)},
  2019.

\bibitem{li2020causal}
Y.~Li, A.~Torralba, A.~Anandkumar, D.~Fox, and A.~Garg, ``{Causal Discovery in
  Physical Systems from Videos},'' \emph{Conf. on Neural Information Processing
  Systems (NeurIPS)}, 2020.

\bibitem{sontakke2020causal}
S.~A. Sontakke, A.~Mehrjou, L.~Itti, and B.~Sch{\"o}lkopf, ``{Causal Curiosity:
  RL Agents Discovering Self-supervised Experiments for Causal Representation
  Learning},'' \emph{arXiv preprint arXiv:2010.03110}, 2020.

\bibitem{pearl2009causality}
J.~Pearl, \emph{Causality}.\hskip 1em plus 0.5em minus 0.4em\relax {Cambridge
  University Press}, 2009.

\bibitem{ahmed2020causalworld}
O.~Ahmed, F.~Tr{\"a}uble, A.~Goyal, A.~Neitz, M.~W{\"u}thrich, Y.~Bengio,
  B.~Sch{\"o}lkopf, and S.~Bauer, ``{CausalWorld: A Robotic Manipulation
  Benchmark for Causal Structure and Transfer Learning},'' \emph{arXiv preprint
  arXiv:2010.04296}, 2020.

\bibitem{devin2018deep}
C.~Devin, P.~Abbeel, T.~Darrell, and S.~Levine, ``{Deep Object-Centric
  Representations for Generalizable Robot Learning},'' \emph{Int'l Conf. on
  Robotics and Automation (ICRA)}, 2018.

\bibitem{yu2017preparing}
W.~Yu, J.~Tan, C.~K. Liu, and G.~Turk, ``{Preparing for the Unknown: Learning a
  Universal Policy with Online System Identification},'' \emph{Robotics:
  Science and Systems (RSS)}, 2017.

\bibitem{nouri2010dimension}
A.~Nouri and M.~L. Littman, ``{Dimension Reduction and its Application to
  Model-Based Exploration in Continuous Spaces},'' \emph{Machine Learning},
  vol.~81, no.~1, pp. 85--98, 2010.

\bibitem{kolter2009regularization}
J.~Z. Kolter and A.~Y. Ng, ``{Regularization and Feature Selection in
  Least-Squares Temporal Difference Learning},'' \emph{Int'l Conf. on Machine
  Learning}, 2009.

\bibitem{parr2008analysis}
R.~Parr, L.~Li, G.~Taylor, C.~Painter-Wakefield, and M.~L. Littman, ``{An
  Analysis of Linear Models, Linear Value-Function Approximation, and Feature
  Selection for Reinforcement Learning},'' \emph{Int'l Conf. on Machine
  Learning}, 2008.

\bibitem{tan2018sim}
J.~Tan, T.~Zhang, E.~Coumans, A.~Iscen, Y.~Bai, D.~Hafner, S.~Bohez, and
  V.~Vanhoucke, ``{Sim-to-Real: Learning Agile Locomotion for Quadruped
  Robots},'' \emph{Robotics: Science and Systems (RSS)}, 2018.

\bibitem{molchanov2019sim}
A.~Molchanov, T.~Chen, W.~Hönig, J.~A. Preiss, N.~Ayanian, and G.~S. Sukhatme,
  ``{Sim-to-(Multi)-Real: Transfer of Low-Level Robust Control Policies to
  Multiple Quadrotors},'' \emph{Int'l Conf. on Intelligent Robots and Systems
  (IROS)}, 2019.

\bibitem{chebotar2019closing}
Y.~Chebotar, A.~Handa, V.~Makoviychuk, M.~Macklin, J.~Issac, N.~Ratliff, and
  D.~Fox, ``{Closing the Sim-to-Real Loop: Adapting Simulation Randomization
  with Real World Experience},'' \emph{Int'l Conf. on Robotics and Automation
  (ICRA)}, 2019.

\bibitem{masson2016reinforcement}
W.~Masson, P.~Ranchod, and G.~Konidaris, ``{Reinforcement Learning with
  Parameterized Actions},'' \emph{AAAI Conf. on Artificial Intelligence}, 2016.

\bibitem{hausknecht2015deep}
M.~Hausknecht and P.~Stone, ``{Deep Reinforcement Learning in Parameterized
  Action Space},'' \emph{Int'l Conf. on Learning Representations (ICLR)}, 2016.

\bibitem{fan2019hybrid}
Z.~Fan, R.~Su, W.~Zhang, and Y.~Yu, ``{Hybrid Actor-Critic Reinforcement
  Learning in Parameterized Action Space},'' \emph{Int'l Joint Conf. on
  Artificial Intelligence (IJCAI)}, 2019.

\bibitem{deisenroth2013survey}
M.~P. Deisenroth, G.~Neumann, J.~Peters, \emph{et~al.}, ``{A Survey on Policy
  Search for Robotics},'' \emph{Foundations and Trends in Robotics}, vol.~2,
  no. 1--2, pp. 1--142, 2013.

\bibitem{battaglia2013simulation}
P.~W. Battaglia, J.~B. Hamrick, and J.~B. Tenenbaum, ``{Simulation as an Engine
  of Physical Scene Understanding},'' \emph{Proceedings of the National Academy
  of Sciences}, vol. 110, no.~45, pp. 18\,327--18\,332, 2013.

\bibitem{ha2018recurrent}
D.~Ha and J.~Schmidhuber, ``{Recurrent World Models Facilitate Policy
  Evolution},'' \emph{Conf. on Neural Information Processing Systems
  (NeurIPS)}, 2018.

\bibitem{scholkopf2021toward}
B.~Sch{\"o}lkopf, F.~Locatello, S.~Bauer, N.~R. Ke, N.~Kalchbrenner, A.~Goyal,
  and Y.~Bengio, ``{Toward Causal Representation Learning},'' \emph{Proceedings
  of the IEEE}, 2021.

\bibitem{peters2010relative}
J.~Peters, K.~M{\"u}lling, and Y.~Altun, ``{Relative Entropy Policy Search},''
  \emph{AAAI Conf. on Artificial Intelligence}, 2010.

\bibitem{konda2000actor}
V.~R. Konda and J.~N. Tsitsiklis, ``{Actor-Critic Algorithms},'' \emph{Conf. on
  Neural Information Processing Systems (NeurIPS)}, 2000.

\bibitem{lu2017expressive}
Z.~Lu, H.~Pu, F.~Wang, Z.~Hu, and L.~Wang, ``{The Expressive Power of Neural
  Networks: A View from the Width},'' \emph{Conf. on Neural Information
  Processing Systems (NeurIPS)}, 2017.

\bibitem{saxe2014exact}
A.~M. Saxe, J.~L. Mcclelland, and S.~Ganguli, ``{Exact Solutions to the
  Nonlinear Dynamics of Learning in Deep Linear Neural Networks},'' \emph{Int'l
  Conf. on Learning Representations (ICLR)}, 2014.

\bibitem{schulman2017proximal}
J.~Schulman, F.~Wolski, P.~Dhariwal, A.~Radford, and O.~Klimov, ``{Proximal
  Policy Optimization Algorithms},'' \emph{arXiv preprint arXiv:1707.06347},
  2017.

\bibitem{stable-baselines}
A.~Hill, A.~Raffin, M.~Ernestus, A.~Gleave, A.~Kanervisto, R.~Traore,
  P.~Dhariwal, C.~Hesse, O.~Klimov, A.~Nichol, M.~Plappert, A.~Radford,
  J.~Schulman, S.~Sidor, and Y.~Wu, ``{Stable Baselines},''
  \url{https://github.com/hill-a/stable-baselines}, 2018.

\bibitem{rusu2016progressive}
A.~A. Rusu, N.~C. Rabinowitz, G.~Desjardins, H.~Soyer, J.~Kirkpatrick,
  K.~Kavukcuoglu, R.~Pascanu, and R.~Hadsell, ``{Progressive Neural
  Networks},'' \emph{arXiv preprint arXiv:1606.04671}, 2016.

\bibitem{Zhou2018}
Q.-Y. Zhou, J.~Park, and V.~Koltun, ``{Open3D: A Modern Library for 3D Data
  Processing},'' \emph{arXiv preprint arXiv:1801.09847}, 2018.

\bibitem{katz2007direct}
S.~Katz, A.~Tal, and R.~Basri, ``{Direct Visibility of Point Sets},'' in
  \emph{ACM SIGGRAPH 2007 papers}, 2007, pp. 24--es.

\bibitem{ester1996density}
M.~Ester, H.-P. Kriegel, J.~Sander, X.~Xu, \emph{et~al.}, ``{A Density-based
  Algorithm for Discovering Clusters in Large Spatial Databases with Noise},''
  in \emph{{Proceedings of the Second Int'l Conf. on Knowledge Discovery and
  Data Mining (KDD)}}, vol.~96, no.~34, 1996, pp. 226--231.

\bibitem{zhang2020modular}
K.~Zhang, M.~Sharma, J.~Liang, and O.~Kroemer, ``{A Modular Robotic Arm Control
  Stack for Research: Franka-Interface and FrankaPy},'' \emph{arXiv preprint
  arXiv:2011.02398}, 2020.

\end{thebibliography}
